\documentclass[review,12pt]{elsarticle}

\usepackage{amssymb}
\usepackage{amsmath}
\usepackage{adjustbox}
\usepackage[numbers]{natbib}
\usepackage{booktabs}
\usepackage[labelfont=bf,labelsep=period]{caption}
\usepackage[hidelinks]{hyperref}

\begin{document}
\begin{frontmatter}



\title{CAST: Cross-Attentive Spatio-Temporal feature fusion for Deepfake detection}


\author[1]{Aryan Thakre\corref{cor1}}
\ead{aryanst21.comp@coeptech.ac.in}


\affiliation[1]{organization={Department of Computer Science and Engineering, COEP Technological University}, 
    city={PUNE},
    state={Maharashtra},
    country={India}}

\author[1]{Omkar Nagwekar}
\ead{nagwekaros21.comp@coeptech.ac.in}

\author[1]{Vedang Talekar}
\ead{talekarvp21.comp@coeptech.ac.in}

\author[1]{Aparna Santra Biswas}
\ead{aparna.comp@coeptech.ac.in}

\cortext[cor1]{Corresponding author}

\begin{abstract}
Deepfakes have emerged as a significant threat to digital media authenticity, increasing the need for advanced detection techniques that can identify subtle and time-dependent manipulations. CNNs are effective at capturing spatial artifacts and Transformers excel at modeling temporal inconsistencies. However, many existing CNN-Transformer models process spatial and temporal features independently. In particular, attention based methods often use independent attention mechanisms for spatial and temporal features and combine them using naive approaches like averaging, addition or concatenation, limiting the depth of spatio-temporal interaction. To address this challenge, we propose a unified CAST model that leverages cross-attention to effectively fuse spatial and temporal features in a more integrated manner. Our approach allows temporal features to dynamically attend to relevant spatial regions, enhancing the model’s ability to detect fine-grained, time-evolving artifacts such as flickering eyes or warped lips. This design enables more precise localization and deeper contextual understanding, leading to improved performance across diverse and challenging scenarios. We evaluate the performance of our model using the FaceForensics++, Celeb-DF, DeepfakeDetection and Deepfake Detection Challenge (DFDC) datasets in both intra and cross dataset settings to affirm the superiority of our approach. Our model achieves strong performance with an Area Under the Curve (AUC) of 99.49\% and an accuracy of 97.57\% in intra-dataset evaluations. In cross-dataset testing, the model achieves AUC scores of 93.31\% and 81.25\% on the unseen DeepFakeDetection and DFDC datasets, respectively. These results highlight the effectiveness of cross-attention-based feature fusion in enhancing the robustness of deepfake video detection.
\end{abstract}


\end{frontmatter}

\section{Introduction}

The rapid advancements in deepfake technologies fueled by advances in generative modeling and deep learning has made it increasingly easy to synthesize realistic facial videos that convincingly mimic real individuals. Such content, while promising for domains like film production, gaming, and virtual reality, poses severe threats to privacy, national security, and the integrity of digital media \cite{roessler2019faceforensicspp}. Consequently, deepfake detection has emerged as a crucial area of research to counter the misuse of these technologies \cite{Celeb_DF_cvpr20}. Existing detection frameworks can be broadly categorized based on their reliance on spatial features, temporal dynamics, or a hybrid of both. Spatial feature-based approaches exploit frame-level artifacts, often introduced during generative synthesis. CNN-based models have been successful in identifying such localized inconsistencies, including texture irregularities and blending artifacts \cite{Li_2020_CVPR}. Recent studies in face-spoof detection \cite {BISWAS1, BISWAS2} demonstrate that spatial cue learning and deep feature representations can also distinguish authentic and manipulated facial content. On the other hand, temporal models capture motion anomalies across frames, which are often introduced due to misalignment in expressions or inconsistencies in facial dynamics. Recurrent networks, 3D CNNs, and Transformer-based methods \cite{10005010,Zheng_2021_ICCV} have shown potential in modeling such spatio-temporal cues. To improve detection robustness, hybrid spatio-temporal models have been proposed. These models typically leverage CNNs for extracting spatial features from each frame and Transformers or RNNs for temporal sequence modeling. In most cases, spatial and temporal branches are treated independently, with feature fusion performed using naive strategies such as concatenation, averaging, or late-stage score fusion \cite{LIU2024128588, 10034609}. This limits the model's capacity to model the intricate interplay between spatial cues and their temporal evolution. Recent works have attempted to address this limitation by integrating attention mechanisms to enhance spatial or temporal features. However, these mechanisms are often applied separately to the spatial and temporal domains with limited cross-interaction. Among these methods,  DFGaze~\cite{10478974} explored spatial-temporal gaze inconsistency, but avoided deep integration of spatial and temporal attention. Similarly, DDL~\cite{10168141} enhanced temporal modeling using an MSA module but avoided deep integration of spatial attention. In recent studies specifically, integrating multi-modal and cross-domain features has shown promise in enhancing generalization to unseen deepfake methods.

While several CNN– and Transformer–based architectures have been explored for deepfake detection, most existing methods still struggle to effectively combine spatial and temporal information in a unified manner. CNNs primarily focus on spatial artifacts such as blending boundaries or texture irregularities, yet they fail to capture frame-level temporal inconsistencies. Conversely, pure Transformer models require extensive data and computation to jointly learn spatial and temporal relations, often leading to limited generalization across unseen manipulations. The core problem this work addresses is the lack of a robust mechanism that can seamlessly fuse spatial and temporal representations while maintaining generalization and resilience to common real-world manipulations. To overcome these limitations, we propose a cross-attention-based spatial–temporal fusion framework (CAST) that enables the temporal stream to selectively attend to spatial cues extracted by the CNN, resulting in more discriminative and robust feature representations for deepfake detection.

Unlike traditional dual-stream approaches, our framework enables temporal tokens derived from transformer outputs to attend to spatial feature tokens extracted from CNNs, fostering rich inter-domain interactions. This dynamic fusion empowers the model to capture complex manipulations, such as flickering eyes or temporally inconsistent facial expressions, which may not be identifiable when modalities are processed in isolation. Our proposed model employs a CNN backbone to extract spatial features from individual frames. These features are then linearly projected and temporally aligned before being passed into a Transformer encoder, which captures temporal dependencies across the video sequence. A multi-head cross-attention module is introduced between the CNN-derived spatial embeddings and Transformer-encoded temporal tokens, enabling effective fusion of local spatial cues and global temporal context. Consequently, the architecture improves generalization across both intra- and cross-manipulation scenarios. Our key contributions can be summarized as follows:

\begin{itemize}
\item We propose a novel cross-attention-based framework for deepfake detection that dynamically fuses spatial and temporal representations for more discriminative video-level embeddings.
\item Our method enables temporal tokens to attend to spatial features, thereby capturing intricate manipulation artifacts that span both domains.
\item We take advantage of multi-head cross-attention to dynamically weigh the importance of different spatial regions over time, which improves the interpretability of the model and focus on key tampered areas.
\item Our framework is designed to be flexible and extensible, making it compatible with various backbone architectures and scalable to larger datasets without significant performance drops.

\end{itemize}

\begin{table}[htbp]
\centering
\caption{Comparison of key deepfake detection methods across categories.}
\label{tab:method_comparison}
\begin{adjustbox}{width=\textwidth}
\begin{tabular}{p{3.3cm}p{0.8cm}p{2.0cm}p{2.6cm}p{2.6cm}p{3.5cm}p{3.5cm}@{}}
\toprule
\textbf{Author} & \textbf{Year} & \textbf{Category} & \textbf{Feature Type} & \textbf{Pre-processing} & \textbf{Strength} & \textbf{Weakness} \\
\midrule

Qiu et al.~\cite{QIU2025103087} & 2025 & Hybrid & Spatial and Frequency features & Bi-directional and Spectral Attention with Feature Superposition & Captures both local spatial and global frequency artifacts & Lacks temporal modeling and relies on handcrafted frequency decomposition \\
\midrule

Wang et al.~\cite{10.1145/3512527.3531415} & 2022 & Hybrid & RGB and Frequency features & Cross Modality Fusion (CMF) block & Captures manipulation cues across spatial and frequency domains & Lacks temporal modeling for video-based detection \\ \midrule

Siddiqui et al.~\cite{SIDDIQUI2025126150} & 2023 & Hybrid & Multi-scale + Attention & CrossViT fusion & Rich cross-modal representation & High model complexity \\ \midrule

Yu et al.~\cite{10138555} & 2023 & Transformer & Multi-view Spatiotemporal & View-wise feature extraction + temporal alignment & Strong cross-dataset generalization & Requires high computational resources \\ \midrule

Wang et al.~\cite{wang2023deepfakeformer} & 2023 & Transformer & Spatio-temporal & Convolutional pooling & Maintains coherence across frames & Prone to overfitting \\ \midrule

Zhao et al.~\cite{Zhao_2021_CVPR} & 2021 & Attention-based & Face regions & Multi-attention heads & Localizes manipulated regions & Sensitive to occlusions \\ \midrule

Face X-ray~\cite{Li_2020_CVPR} & 2020 & Spatial & Blending artifacts & Region map extraction & Fine-grained boundary detection & Misses global inconsistencies \\ \midrule

Yoon et al.~\cite{YOON2024102424} & 2024 & Multimodal & Facial, Audio, Text cues & Triple-modality fusion & Zero-shot generalization to unseen identities & Higher computational cost \\ \midrule

Chen et al.~\cite{Chen_2022_CVPR} & 2022 & Self-supervised & Adversarial features & No labels needed & Generalizes to unseen fakes & May overfit synthetic noise \\ \midrule

Hsu et al.~\cite{10798466} & 2025 & Diffusion / Inpainting & Tiny partial fingerprint synthesis & Keypoint-guided inpainting diffusion & Generates 818K synthetic samples, 99.1\% accuracy, strong augmentation & Domain-specific to fingerprints, limited applicability to video modalities \\ 

\bottomrule
\end{tabular}
\end{adjustbox}
\end{table}

\section{Related Work}
In this section, we discuss about some key developments that have occured in the field of deepfake videos. We categorize prior works in two types: 

\begin{itemize}
    \item 
    Deepfake generation: Certain methods used to synthesize high-fidelity facial manipulations.
    \item
    Deepfake detection: Some recent approaches designed to identify such forgeries using various spatial, temporal, and multimodal cues.
\end{itemize}

\subsection{Deepfake Generation}

Deepfake generation refers to the use of deep learning models to synthesize manipulated facial videos that are visually indistinguishable from authentic ones. These techniques often involve replacing or altering the facial identity, expressions, or movements of a source individual to match those of a target subject. Early methods employed different studies based on autoencoder-decoder architectures used in the DeepFake framework~\cite{Korshunova_2017_ICCV}, where separate encoders and decoders are trained for two identities, allowing for face-swapping via a shared latent space. Models like FaceSwap and DeepFakeEncoder relied on identity-preserving autoencoders, while more advanced techniques such as Face2Face and Deep Video Portraits~\cite{10.1145/3306346.3323035} enabled fine-grained facial reenactment and head pose manipulation in real-time. GAN-based systems like Neural Textures~\cite{10.1145/3306346.3323035} produced highly detailed and consistent outputs by learning rich texture representations. Understanding these generative frameworks is crucial for developing effective detection strategies. The continued progression of these generative techniques, especially with transformer-based generators, poses an escalating challenge for detection methods, as artifacts become increasingly subtle and temporally stable.

Early deepfake generation methods were primarily based on autoencoders and classical GAN models such as DeepFake-Autoencoder~\cite{deepfake_autoencoder}, enabling identity reconstruction from paired face images. With the introduction of advanced GAN architectures, particularly StyleGAN and StyleGAN2~\cite{karras2019stylegan, karras2020stylegan2}, synthesis quality dramatically improved by enabling high-resolution texture control and realistic facial structure manipulation. StyleGAN3~\cite{karras2021stylegan3} later addressed temporal instability by eliminating aliasing artifacts, making generated video content visually more coherent.

Recently, diffusion-based generative models, such as DDPM~\cite{ho2020ddpm} and Latent Diffusion~\cite{rombach2022latent}, have demonstrated superior realism and semantic controllability compared to GANs through iterative denoising processes. These approaches have been increasingly adopted for face swapping and reenactment tasks under unconstrained conditions. Alongside diffusion models, transformer-based architectures have gained attention for deepfake generation, leveraging global self-attention to capture long-range dependencies more effectively than CNN-based designs. Vision Transformer-based generation frameworks~\cite{dosovitskiy2021vit} and masked autoencoder-based methods enable improved temporal coherence and semantic feature modeling across video frames.

In addition, multimodal deepfake generation frameworks combining face synthesis with speech and 3D modeling have emerged. Audio-driven reenactment models such as Wav2Lip~\cite{prajwal2020wav2lip} and SadTalker~\cite{zhang2023sadtalker} generate realistic lip synchronization, while 3D NeRF-based reenactment approaches~\cite{nerf2020} produce high-fidelity facial dynamics and viewpoint control. Overall, the rapid evolution from autoencoders to GAN, diffusion, transformer, and multimodal frameworks has significantly increased the realism and complexity of manipulated media, reinforcing the necessity for robust detection models such as the proposed CAST framework.

\subsection{Deepfake Detection}
Early works relied on spatial cues using CNNs like XceptionNet~\cite{Chollet_2017_CVPR}, which achieved high performance on FaceForensics++~\cite{roessler2019faceforensicspp}. Later hybrid models used CNN--LSTM or CNN--Transformer designs to capture spatio-temporal features. Temporal modeling using Vision Transformers has seen continued evolution. Recent work such as MSVT~\cite{10138555} proposed a transformer that captures multiple spatiotemporal views, allowing for robust generalization across varied manipulation methods.. With the rapid growth of deepfake generation techniques, a corresponding surge in deepfake detection methods has emerged to counter manipulated media, which is summarized in Table \ref{tab:method_comparison}. These detection methods vary across spatial, temporal, spatio-temporal, multimodal, and attention-based approaches. Biswas et al.~\cite{BISWAS1} utilized compact Xception framework with convolution block attention module to extract illumination invariant, chromaticity and depth features for the prediction of real and fake images. Rössler et al.~\cite{roessler2019faceforensicspp} created the widely used FaceForensics++ dataset, providing a standardized benchmark for evaluating manipulated facial videos. Extending this, ForgeryNet~\cite{He_2021_CVPR} introduced a more diverse and large-scale dataset to facilitate robust training and cross-domain evaluation. These datasets helped reveal generalization issues in early detectors, especially when models trained on one manipulation type failed to detect others effectively. To improve generalizability, Li et al.~\cite{Li_2020_CVPR} developed Face X-ray, a method capable of identifying blending artifacts between manipulated and authentic facial regions. This was extended by Zhao et al.~\cite{Zhao_2021_CVPR} who used multi-attentional mechanisms to highlight discriminative facial regions. Siddiqui et al.~\cite{SIDDIQUI2025126150} enhanced this approach using DenseNet and CrossViT to fuse multi-scale features and attention-based cues for robust classification. Wang et al.~\cite{10.1145/3512527.3531415} proposed M2TR, which employs a Cross Modality Fusion (CMF) block to perform cross-attention based fusion between RGB and frequency features. Soudy et al.~\cite{Soudy2024} introduced a hybrid deepfake detection approach that combines CNNs with a Vision Transformer, organizing the process into intuitive stages of preprocessing, targeted feature analysis, and final prediction through majority voting. Qiu et al.~\cite{QIU2025103087} proposed D2-Fusion, which combines bi-directional spatial attention and spectral frequency attention, followed by a feature superposition strategy to amplify artifact distinctions. Similarly, Wang et al.~\cite{wang2023deepfakeformer} adopted a convolutional pooling transformer to further enhance spatio-temporal coherence understanding in video forgeries. Multimodal detection has also gained popularity due to its ability to analyze non-visual inconsistencies. Yoon et al.~\cite{YOON2024102424} proposed TMI-former, a triple-modality deepfake detection framework that exploits facial, audio, and text modality interactions to detect forgeries in a zero-shot setting. Choi et al.~\cite{Choi_2024_CVPR} proposed a hybrid model combining a 3D ResNet with a StyleGRU module to capture both spatio-temporal features and latent style dynamics, improving deepfake detection generalization. Chen et al.~\cite{CHEN2023109179} proposed a DeepFake detection framework based on Bi-Granularity Artifacts, capturing both intrinsic (e.g., upsampling) and extrinsic (e.g., blending) forgery traces. Their multi-task learning approach effectively exploits these artifacts to improve detection across diverse datasets. Tolosana et al.~\cite{tolosana2020deepfakes} also reviewed the broad landscape of face manipulation and fake detection, highlighting persistent challenges such as cross-domain generalization, real-time detection, and deepfake localization. Recent advancements have further enhanced the robustness of deepfake detection through hybrid and self-supervised approaches. Chen et al.~\cite{Chen_2022_CVPR} introduced a self-supervised learning method that leveraged adversarial examples for improved generalization ability of detection models and enabled better performance across varied manipulation types. Sun et al.~\cite{Sun_Yao_Chen_Ding_Li_Ji_2022} proposed a dual contrastive learning framework that aligns intra and inter-modality features using instance- and prototype-level losses, enhancing generalization under domain shifts. Zheng et al.~\cite{Zheng_2021_ICCV} proposed a temporal coherence-based strategy to address the limitations of static frame analysis by enforcing consistency across sequential frames, thus improving the detection of subtle and temporally coherent forgeries. Hsu et al.~\cite{10798466} proposed an Inpainting Diffusion Synthetic approach guided with feature keypoints for generating tiny partial fingerprint samples, producing over 818,077 synthetic images and achieving up to 99.1\% matching accuracy. Their work demonstrates the significance of high-quality data augmentation for improving recognition robustness.

These studies collectively illustrate the evolution of deepfake detection from early handcrafted or shallow CNN-based methods to complex spatio-temporal, attention-based, and multimodal architectures. While performance continues to improve, key challenges remain in terms of generalization across unseen manipulation types, ineffective fusion strategies and detection in real-time streaming environments.

\section{Method}
\begin{figure*}[h]
\centering
\includegraphics[scale=0.13]{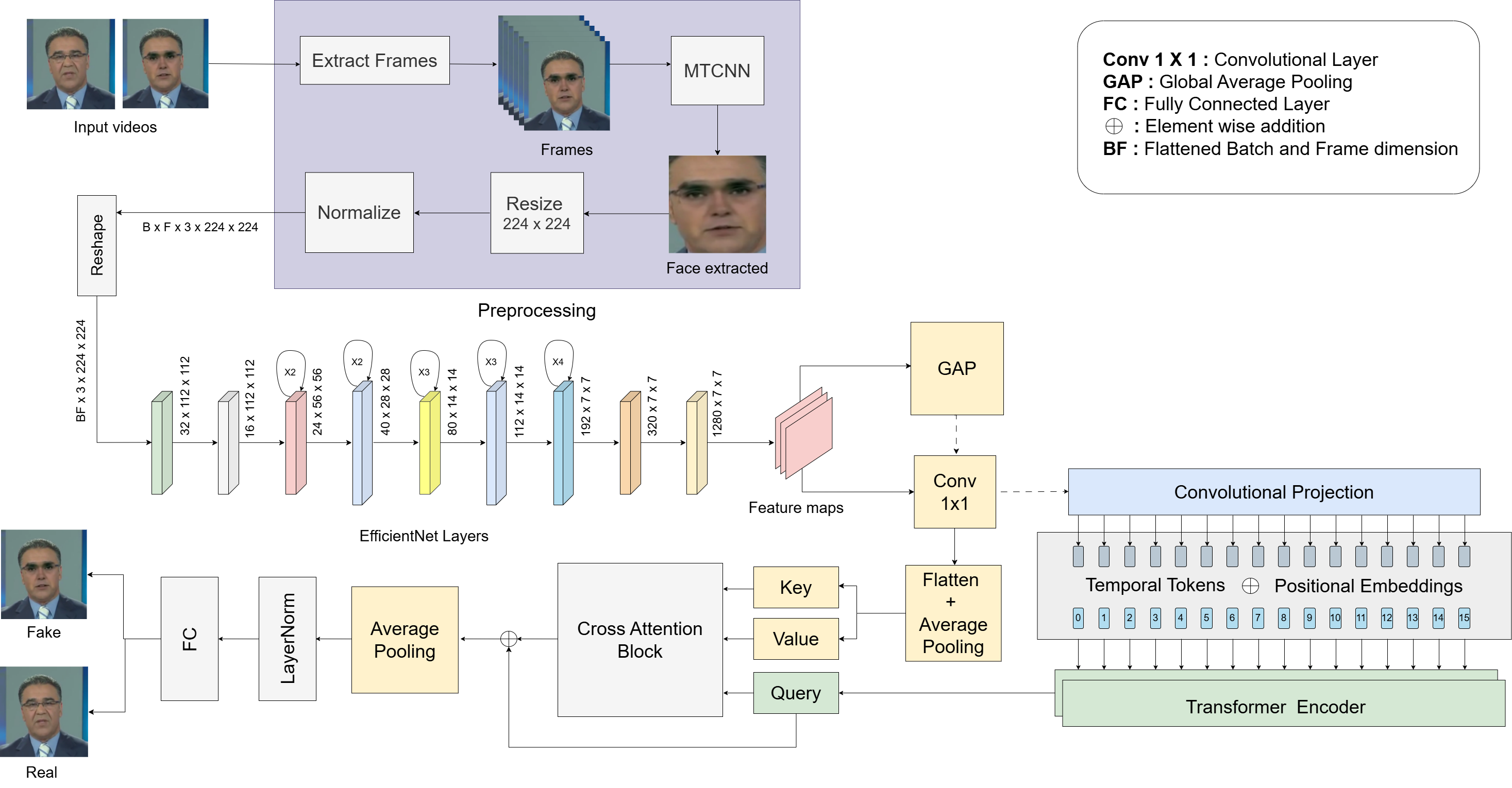}
\caption{The pipeline of our proposed framework CAST which consists of an EfficientNet-Transformer hybrid architecture with cross-attention based spatio-temporal fusion.}
\label{fig:pipeline}
\end{figure*}

In this section, we present the complete architecture employed in our framework as shown in Fig.~\ref{fig:pipeline}. Initially, video samples are subjected to preprocessing, wherein frames are extracted and aligned using the Multi-Task Cascaded Convolutional Network~\cite{zhang2016joint}. This preprocessing step ensures spatial consistency and prepares the input for subsequent feature extraction. We consider a fixed number of 16 frames from each video during the training and testing phases of our model, which are uniformly sampled to capture temporal diversity.

\begin{figure}[h]
\centering
\includegraphics[scale=0.20]{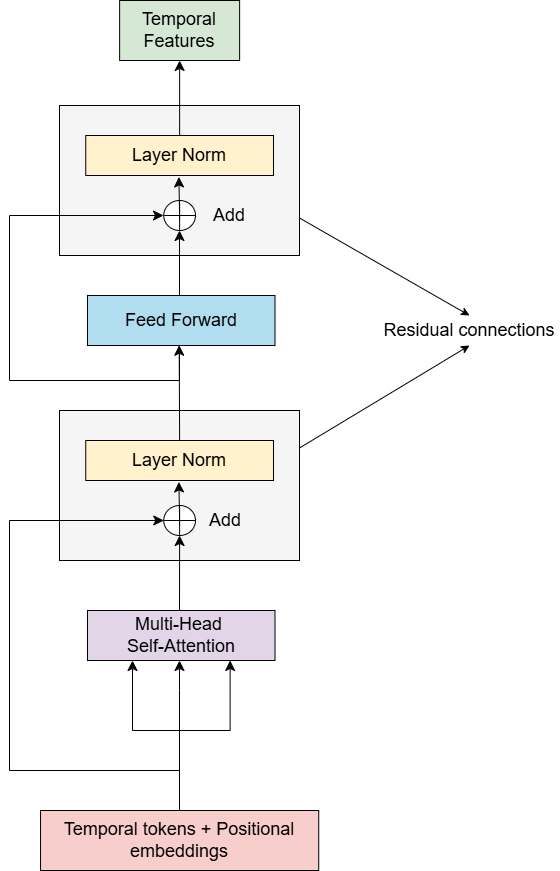}
\caption{The working of Transformer encoder (single head) modeling temporal dependencies using features extracted from EfficentNet layers}
\label{fig:transformer_encoder}
\end{figure}

\begin{figure*}[h]
\centering
\includegraphics[scale=0.20]{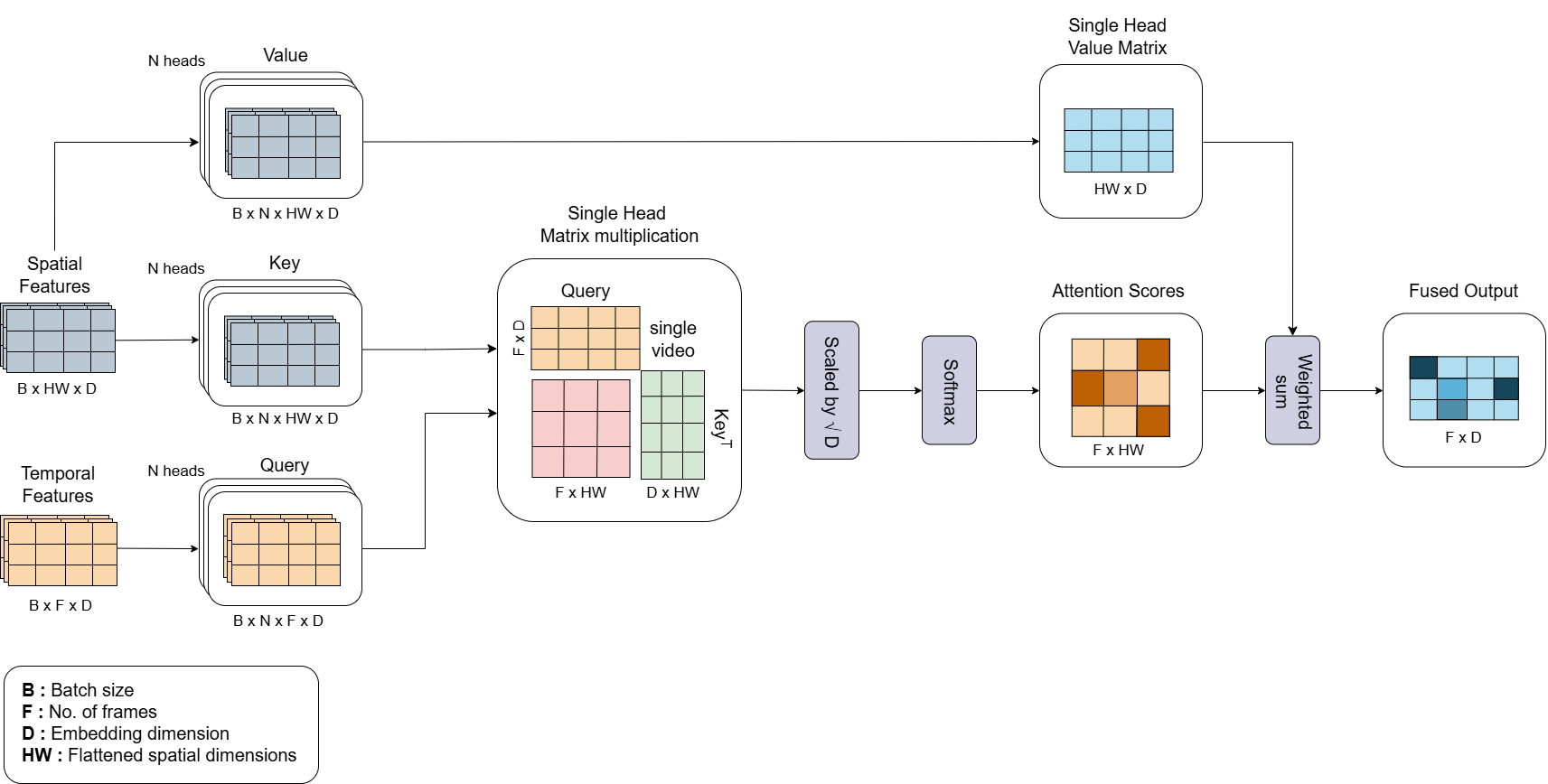}
\caption{Working of Cross-Attention module in the CAST Model.}
\label{fig:cross_attention}
\end{figure*}

\subsection{Frame Selection and Preprocessing}
\label{sec:frame_selection}

Given an input video $\mathcal{V}$ characterized by a native frame rate $f_{\text{orig}}$ (frames per second), we aim to ensure temporal consistency across diverse video samples by extracting a fixed number of semantically meaningful frames. To do so, we define a target sampling rate $r$, and calculate the inter-frame interval as:
\begin{equation}
\Delta = \max\left(1, \left\lfloor \frac{f_{\text{orig}}}{r} \right\rfloor \right)
\end{equation}
where $\Delta$ denotes the frame sampling interval. This ensures that frames are temporally spaced to capture diverse visual dynamics while maintaining computational feasibility.

For a video of length $L$ frames, we extract $F$ frames using:
\begin{equation}
F = \left\lfloor \frac{L}{\Delta} \right\rfloor.
\end{equation}

Each selected frame $I_i \in \mathbb{R}^{H \times W \times 3}$ is then processed through a face detection and alignment step using the Multi-task Cascaded Convolutional Network (MTCNN) \cite{zhang2016joint}, which ensures consistent facial region cropping and alignment across varying poses and lighting conditions. The detected face regions constitute the regions of interest (ROI) fed to the downstream classification network.

\subsubsection{Spatial Resizing and Normalization}
Each detected face crop $F_i$ is resized to a fixed spatial resolution of $224 \times 224$ pixels to match the input requirements of modern CNN backbones. To ensure statistical consistency with ImageNet-pretrained feature extractors, we apply per-channel normalization:
\begin{equation}
\tilde{F}_i = \frac{F_i - \mu}{\sigma}, \quad \mu = [0.485, 0.456, 0.406], \quad \sigma = [0.229, 0.224, 0.225],
\end{equation}
where $\mu$ and $\sigma$ denote the empirical mean and standard deviation of RGB channels in the ImageNet dataset~\cite{5206848}. This transformation centers the data and stabilizes the learning dynamics of deep CNNs.

The final preprocessed batch of frame-wise ordered and spatially normalized face crops is represented as:
\begin{equation}
\mathcal{X} = \left\{ \tilde{F}i \in \mathbb{R}^{3 \times 224 \times 224} \right\}_{i=1}^{F}.
\end{equation}

This sequence $\mathcal{X}$ serves as the primary input to the CNN backbone.

\subsection{Feature Extraction via CNN Backbone}

Each normalized face image $\tilde{F}_i$ is passed through the EfficientNet~\cite{pmlr-v97-tan19a} backbone to extract spatial feature representations:
\begin{align}
M_i &= \text{CNN}(\tilde{F}_i) \\
    &= f_{\text{CNN}}(\tilde{F}_i; \theta_{\text{CNN}}) \in \mathbb{R}^{C \times H' \times W'}
\end{align}
where $f_{\text{CNN}}(\cdot)$ denotes the CNN feature extractor parameterized by $\theta_{\text{CNN}}$, and $M_i$ is the resulting feature map with $C$ channels and spatial dimensions $H'\times W'$. These intermediate features capture multi-scale semantic information across varying receptive fields, enabling the detection of fine-grained manipulation artifacts.

To prepare spatial tokens for the cross-attention module, we apply a $1\times1$ pointwise convolution to project the feature maps from $C$ to a lower-dimensional space:
\begin{align}
E_i &= \text{Conv}_{1 \times 1}(M_i) \\
    &= W_c * M_i + b_c \in \mathbb{R}^{d \times H' \times W'}
\end{align}
where $W_c \in \mathbb{R}^{d \times C}$ and $b_c \in \mathbb{R}^{d}$ are learnable projection parameters. This pointwise convolution plays a critical role in aligning the channel dimensions of the CNN-derived features with the downstream modules. By reducing the original $C$-dimensional features to a more compact $d$-dimensional representation, it not only ensures compatibility with the cross-attention mechanism but also improves computational efficiency. Moreover, the $1\times1$ convolution acts as a lightweight feature selection layer, emphasizing informative semantic patterns while suppressing irrelevant activations, thereby facilitating more focused spatial attention.
 The resulting tensor is reshaped into a sequence of spatial tokens:
\begin{align*}
S_i &= \text{reshape}(E_i) \\
    &= [E_i^{(1)}, \ldots, E_i^{(H'W')}] \in \mathbb{R}^{H'W' \times d}
\end{align*}
where each token $E_i^{(j)}$ represents a localized semantic embedding corresponding to a patch in the original face image. These spatial tokens serve as the key-value inputs in the cross-attention module, allowing the temporal representation to selectively attend to spatial regions.

\subsection{Temporal Token Computation and Encoding}

To derive temporal tokens from the CNN features extracted for each frame, we first apply Global Average Pooling (GAP) to summarize the spatial information:
\begin{equation}
\text{GAP}(M_i) = \frac{1}{H'W'} \sum_{h=1}^{H'} \sum_{w=1}^{W'} M_i[:, h, w] \in \mathbb{R}^C
\end{equation}
Here, $M_i \in \mathbb{R}^{C \times H' \times W'}$ is the CNN feature map of the $i^{\text{th}}$ frame. GAP condenses this spatial map into a compact $C$-dimensional vector by averaging over all spatial positions, preserving global context and reducing dimensionality. Next, this pooled representation is projected into a latent embedding space using a $1 \times 1$ convolution:
\begin{align}
T_i &= \text{Conv}_{1 \times 1}(\text{GAP}(M_i)) \in \mathbb{R}^d \\
    &= W_t \cdot \text{GAP}(M_i) + b_t \in \mathbb{R}^d
\end{align}
Here, $W_t \in \mathbb{R}^{d \times C}$ and $b_t \in \mathbb{R}^d$ are the learnable weights and bias of the convolutional projection layer. This operation maps the $C$-dimensional feature into a fixed $d$-dimensional embedding, which we refer to as the temporal token $T_i$. The full sequence of temporal tokens across $F$ frames is then organized as:
\begin{equation}
\mathcal{T} = [T_1, T_2, \ldots, T_F] \in \mathbb{R}^{F \times d}
\end{equation}
The per-frame features $T_i$ are stacked into a sequence $\mathcal{T} \in \mathbb{R}^{F \times d}$, where $F$ denotes the number of frames and $d$ is the embedding dimension. This sequence $\mathcal{T}$ serves as an ordered representation of the video, where each row corresponds to a frame-level token embedding.

We used a transformer encoder to compute temporal relationships between frames and detect motion-related inconsistencies. While the convolutional backbone focuses on extracting spatial features within each frame, it lacks the ability to reason about how these features evolve over time. The transformer addresses this limitation by operating on the sequence of frame-level representations and learning dependencies across the temporal dimension. To incorporate temporal ordering, we add learnable positional embeddings $\mathcal{P} \in \mathbb{R}^{F \times d}$. This enables the model to distinguish token order and effectively capture temporal dynamics across frames:
\begin{equation}
\mathcal{T}_{\text{pos}} = \mathcal{T} + \mathcal{P}
\end{equation}

The enriched sequence $\mathcal{T}_{\text{pos}}$ is fed into the Transformer encoder:
\begin{equation}
\mathcal{Z} = \text{TransformerEncoder}(\mathcal{T}_{\text{pos}})
\end{equation}

Fig.~\ref{fig:transformer_encoder} depicts the internal structure of a single head Transformer encoder. It consists of a multi-head attention layer, feed-forward networks, and residual connections, all interleaved with layer normalization. The core component responsible for capturing long range dependencies between frames is the Multi-Head Self-Attention (MHSA) module, which allows each temporal token to attend to all others in the sequence.

Through multiple layers of attention and non-linear transformations, the Transformer builds a temporally-aware representation $\mathcal{Z} \in \mathbb{R}^{F \times d}$ that captures how spatial features transition across time. This is particularly important in tasks like deepfake detection, where subtle temporal inconsistencies, such as flickering artifacts, unnatural motion, or identity mismatches across frames, can serve as key indicators of manipulation. By modeling such temporal dynamics, the Transformer enhances the model's ability to identify such inconsistencies effectively.

\subsection{Feature Fusion via Cross-Attention}

To obtain a stable and representative spatial context for the entire video, we compute a mean of the spatial tokens across all frames:
\begin{equation}
    S_{\text{mean}} = \frac{1}{F} \sum_{i=1}^{F} S_i
\end{equation}
Here, $S_i$ denotes the set of spatial tokens extracted from the $i^\text{th}$ frame. Averaging across the temporal dimension smooths out transient spatial variations while preserving the dominant spatial features consistent throughout the video. This aggregated spatial token set, $S_{\text{mean}}$, serves as the key and value in the subsequent cross-attention mechanism, enabling each temporal token to selectively attend to spatial cues during the fusion stage.

A cross-attention mechanism is applied to fuse the spatial tokens $S_{\text{mean}}$ with the temporal tokens $\mathcal{Z}$ :
\begin{equation}
    \hat{\mathcal{Z}} = \text{softmax}\left( \frac{\mathcal{Z}W_Q (S_{\text{mean}}W_K)^\top}{\sqrt{d}} \right)(S_{\text{mean}}W_V)
\end{equation}
\vspace{-12mm}
\begin{equation}
    \mathcal{F}_{\text{fused}} = \text{LayerNorm}(\mathcal{Z} + \hat{\mathcal{Z}})
\end{equation}

Here, \( \mathcal{Z}W_Q \), \( S_{\text{mean}}W_K \), and \( S_{\text{mean}}W_V \) represent the query, key, and value projections, respectively. As shown in Fig.~\ref{fig:cross_attention}, the cross-attention mechanism functions by projecting the input sequences into three distinct representations: queries ($Q$), keys ($K$), and values ($V$). Each query vector $q_i \in Q$ retrieves relevant information by computing similarity with key vectors $k_j \in K$. The attention weights are calculated using scaled dot-product attention:

\vspace{-7mm}
\begin{equation}
    A = \text{softmax}\left( \frac{QK^\top}{\sqrt{d}} \right)
\end{equation}

The attention weights \( A \), shown as the orange grid in Fig.~\ref{fig:cross_attention}, indicate the relevance of each key to a given query. The output, represented as the blue grid, is obtained by computing a weighted sum over the values:
\begin{equation}
    \hat{z}_i = \sum_{j=1}^{n_k} A_{ij} \cdot V_j
\end{equation}
where \( \hat{z}_i \) is the output corresponding to the \( i^{\text{th}} \) query, \( A_{ij} \) is the attention weight between the \( i^{\text{th}} \) query and \( j^{\text{th}} \) key, and \( V_j \) is the value vector associated with the \( j^{\text{th}} \) key.

This formulation allows each query to selectively attend to and integrate contextual information from the entire set of keys and values

In our case, the temporal tokens $\mathcal{Z}$ serve as queries ($Q$), and the pooled spatial tokens $S_{\text{mean}}$ act as both keys ($K$) and values ($V$), enriching temporal features with spatial cues for more informed representation learning. During cross-attention, each temporal token in \( \mathcal{Z} \) computes a relevance score with every spatial token in \( S_{\text{mean}} \) through scaled dot-product similarity. This results in an attention matrix \( A \), where each entry \( A_{ij} \) quantifies how strongly the \( i^{\text{th}} \) temporal token attends to the \( j^{\text{th}} \) spatial token. These attention weights are then used to aggregate a weighted sum over the spatial value vectors \( V_j \), allowing temporal tokens to integrate spatial context that is most relevant to their positions. The fused representation $\mathcal{F}_{\text{fused}}$ thus carries both temporal dependencies and spatial coherence, facilitating a more comprehensive and discriminative feature space for downstream classification tasks.

In the proposed cross-attention fusion, temporal tokens are used as \textit{queries} while spatial tokens operate as \textit{keys} and \textit{values}. This formulation reflects the functional relationship between temporal and spatial information in deepfake analysis. Temporal dynamics frequently expose inconsistencies such as motion discontinuities, lip-sync mismatch, or frame-level blending artifacts, which are strong indicators of manipulation. By treating temporal features as queries, the model effectively asks: \textit{“Which spatial regions are relevant given the current temporal context?”} In contrast, spatial features represent a more stable repository of structural appearance cues, functioning as a spatial memory from which values are selectively retrieved based on learned relevance. The attention weights compute similarity between temporal queries and spatial keys, and the resulting value aggregation enhances spatial details precisely where temporal cues detect anomalies. Empirically, our directionality ablation study (Sec. 5.5.3, Table 6) demonstrates that reversing the Q/K roles leads to reduced cross-dataset AUC performance, supporting this asymmetric design choice.

\subsection{Classification Layer and Optimization}

The video-level representation is obtained by averaging all frame-level tokens:
\begin{equation}
    \bar{z} = \frac{1}{F} \sum_{i=1}^{F} \mathcal{F}_{\text{fused}}^{(i)}
\end{equation}

This vector is passed through a linear classifier to produce a logit score:
\begin{equation}
    \hat{z} = W \bar{z} + b
\end{equation}

The computed logit is passed to the binary cross-entropy loss with logits for calculating the loss value.
The binary cross-entropy loss with logits for a label $y \in \{0, 1\}$ works as follows:
\begin{equation}
    \mathcal{L}_{\text{BCE}} = \log(1 + e^{-\hat{z}}) + (1 - y)\hat{z}
\end{equation}

This formulation avoids numerical instability by combining the sigmoid and log operations into a single, stable function. Parameters $\theta$ are updated using gradient descent with scaling for mixed-precision training:
\begin{equation}
    \theta \leftarrow \theta - \eta \cdot \text{scale}(\nabla_\theta \mathcal{L}_{\text{BCE}})
\end{equation}

\section{Experiments}

In this section, we present the experimental setup used to evaluate the performance of our proposed model CAST. We first describe the publicly available benchmark datasets used in our study, followed by the training configurations and optimization strategies. Finally, we outline the evaluation metrics used to assess the performance and generalization capability of our model across intra- and cross-dataset scenarios.

\subsection{Datasets}

To comprehensively evaluate our proposed cross-attentive CNN-Transformer model for deepfake video detection, we conducted extensive experiments on three publicly available benchmark datasets: FaceForensics++~\cite{roessler2019faceforensicspp}, DeepfakeDetection~\cite{DDD_GoogleJigSaw2019}, Celeb-DF (v2)~\cite{Celeb_DF_cvpr20} and Deepfake Detection Challenge~\cite{DFDC2020}. These datasets encompass a wide range of manipulation techniques and visual qualities, enabling robust assessment across both intra- and cross-dataset scenarios.

\textbf{FaceForensics++ (FF++):} We utilized the high-quality (c23) compressed version of FF++, which contains videos manipulated using four different techniques: Deepfakes (DF), Face2Face (F2F), FaceSwap (FS), and NeuralTextures (NT). For each manipulation method, we adopted the official split provided by the dataset: 720 manipulated and 720 pristine videos for training, 140 manipulated and 140 pristine videos for validation, and another 140 manipulated and 140 pristine videos for testing. We trained separate models for each manipulation type to assess both intra-manipulation performance (testing on the same type) and cross-manipulation generalization (testing on the remaining types).

\textbf{DeepfakeDetection (DFD):} The DeepfakeDetection (DFD) dataset was released by Google and consists of over 3,000 high-resolution videos featuring both real and synthetically manipulated faces. These videos were created using multiple deepfake generation methods applied across various subjects, environments, and lighting conditions. DFD serves as a valuable benchmark for evaluating deepfake detection systems due to its realistic manipulations and diversity in terms of demographics, pose, and expressions. In our work, we utilized a curated subset of pristine and fake videos from DFD to conduct cross-dataset evaluations. The dataset's diversity and naturalistic quality make it a robust benchmark for assessing generalization capability of detection models.

\textbf{Celeb-DF (v2):} The Celeb-DF (v2) is a challenging deepfake dataset that contains over 5,600 videos generated from more than 590 real source videos of celebrities. The deepfakes in this dataset were created using an improved synthesis algorithm that significantly reduces visual artifacts, such as flickering or low texture quality, often found in earlier datasets. As a result, Celeb-DF v2 provides a more realistic and difficult benchmark for deepfake detection. Its subtle manipulations and high-quality video characteristics help assess a model's ability to generalize to unseen and less detectable forgeries. In our evaluation pipeline, Celeb-DF (v2) is primarily used for cross-dataset testing to benchmark the generalization performance of our proposed architecture.

\textbf{Deepfake Detection Challenge (DFDC)}: The Deepfake Detection Challenge (DFDC) dataset is a large-scale benchmark for deepfake video detection, comprising of 19,154 real videos and 100,000 manipulated videos generated from 3,426 compensated actors under controlled and unconstrained recording conditions. The manipulated videos were created using multiple deepfake generation pipelines, resulting in diverse visual artifacts, identity substitutions, and temporal inconsistencies. The dataset exhibits substantial variability in facial attributes, head poses, illumination conditions, and background scenes, making it particularly challenging for deepfake detection. It introduces significant distributional shifts and weak manipulation cues that closely resemble real-world forgery scenarios. In this work, DFDC is primarily used for cross-dataset evaluation to assess the generalization capability of the proposed model across diverse data distributions and manipulation patterns.

\subsection{Experimental settings}

All experiments and model development are conducted on two NVIDIA T4 GPUs. For backbone configurations with higher computational requirements such as EfficientNet-B5, we utilized PyTorch’s \texttt{DataParallel} wrapper to distribute model computations across both GPUs. This allowed us to significantly reduce per-epoch training time and utilize available hardware efficiently. To further optimize memory usage and computational throughput, we incorporated mixed-precision training using the \texttt{torch.cuda.amp} module. Specifically, we employed dynamic loss scaling via the \texttt{GradScaler} utility, which helps maintain training stability by preventing underflow during gradient computation. The forward and backward passes were enclosed within \texttt{autocast()} contexts to ensure that appropriate operations were executed in lower precision (FP16), while sensitive operations such as gradient accumulation remained in full precision (FP32). Training was performed using the PyTorch framework. We used the Adam optimizer with an initial learning rate of $1 \times 10^{-4}$ and a weight decay of $1 \times 10^{-5}$. A batch size of 8 video clips are employed across all experiments, constrained by GPU memory and batch-level padding for uniformity. We have used the loss function binary cross-entropy with logits, and a dropout rate of 0.3 is applied consistently across modules to mitigate overfitting. Each model variant is trained for a maximum of 25 epochs. Instead of employing early stopping, we adopted an early model saving strategy, where the model checkpoint corresponding to the lowest validation loss during training was preserved for final evaluation. This approach ensured that the reported results reflected the most optimal model performance.
As we conducted experiments across multiple CNN backbones, training times varied depending on the backbone used. EfficientNet-B5 and XceptionNet, required significantly more time per epoch compared to EfficientNet-B0. However, EfficientNet-B5 gained proportionally more advantage from the dual-GPU configuration and precision scaling.

\subsection{Evaluation Metrics:} To quantitatively assess the performance of our proposed deepfake detection model, we utilize several standard binary classification metrics, including Accuracy and Area Under the Curve (AUC). These metrics are computed based on the predicted labels for each video, where each video-level prediction is obtained by averaging the model’s sigmoid outputs across the 16 extracted frames and applying a threshold of 0.5.

Let True Positives ($TP$) represent the number of fake videos correctly classified as fake, True Negatives ($TN$) the number of real videos correctly classified as real, False Positives ($FP$) the number of real videos incorrectly classified as fake, and False Negatives ($FN$) the number of fake videos incorrectly classified as real. The evaluation metrics are formulated as follows:
\begin{equation}
\text{Accuracy} = \frac{TP + TN}{TP + TN + FP + FN}
\end{equation}

\begin{equation}
\text{AUC} = \int_{0}^{1} TPR(FPR) \, dFPR
\end{equation}

Here, AUC refers to the area under the Receiver Operating Characteristic (ROC) curve, which plots the True Positive Rate (TPR) against the False Positive Rate (FPR) at various threshold settings. A higher AUC indicates better model performance in distinguishing between real and fake videos across different classification thresholds.

\section{Experimental results and discussion:}

To comprehensively evaluate the performance and generalization capability of our proposed method, we conduct experiments across four key evaluation protocols: single-manipulation evaluation, intra-dataset testing, cross-dataset generalization, and multi-source manipulation assessment. These evaluations are performed primarily on the
FaceForensics++~\cite{roessler2019faceforensicspp} dataset under high-quality compression settings, along with DFD~\cite{DDD_GoogleJigSaw2019}, DFDC~\cite{DFDC2020} and Celeb-DF (v2)~\cite{Celeb_DF_cvpr20} datasets. We adopt standard classification metrics, including AUC and Accuracy, to ensure a consistent and fair comparison with existing state-of-the-art (SOTA) approaches.

\subsection{Single Manipulation Evaluation}

We evaluate our proposed CNN-Transformer hybrid architecture with cross-attention fusion on the FF++(HQ) dataset under the single-manipulation training setting. Each model is trained using one specific manipulation type: Deepfakes (DF), Face2Face (F2F), FaceSwap (FS), or NeuralTextures (NT) and evaluated on all four manipulation types. The AUC scores are reported in Table~\ref{tab:single_manipulation_auc}, where we compare our models using EfficientNet-B0 (CAST-B0) and EfficientNet-B5 (CAST-B5) backbones with several state-of-the-art methods: D\textsuperscript{2}-Fusion~\cite{QIU2025103087}, DCL~\cite{Sun_Yao_Chen_Ding_Li_Ji_2022}, RECCE~\cite{Cao_2022_CVPR}, ADAL~\cite{9931753}, and WATCHER~\cite{Wang2024}.

When trained on Deepfakes (DF), our CAST-B0 model achieves a perfect AUC of 100.00\% on DF, and the highest average AUC (84.65\%) across all test sets. It also achieves the best AUC on F2F (86.29\%) and NT (92.02\%), outperforming strong baselines such as ADAL (83.65\%) and WATCHER (84.58\%). This indicates that CAST-B0 can effectively learn generalized features from the DF manipulation type, which contains rich spatial and temporal artifacts. Under the Face2Face (F2F) training setting, CAST-B0 again achieves the highest average AUC of 83.79\%, demonstrating strong generalization to DF (94.33\%) and NT (82.74\%). Although ADAL slightly outperforms our model on FS (69.49\% vs. 58.42\%), CAST-B0 remains competitive across all other test cases. This suggests that training on F2F provides a diverse and transferable feature space, which our cross-attention architecture is able to leverage effectively. In contrast, performance drops are observed when training on FaceSwap (FS). Here, our models (CAST-B0: 73.44\%, CAST-B5: 71.61\% average AUC) underperform compared to D\textsuperscript{2}-Fusion (76.41\%) and WATCHER (79.58\%). Despite nearly perfect performance on FS itself (CAST-B0: 99.29\%, CAST-B5: 99.99\%), generalization to DF (59.97\%) and NT (51.11\%) is limited. This suggests that FS introduces relatively weaker temporal inconsistencies and less diverse artifacts, which hampers the ability of our model to learn cross-manipulation representations via attention-based mechanisms. Finally, when trained on NeuralTextures (NT), both our models perform strongly on NT itself (CAST-B0: 97.82\%, CAST-B5: 97.97\%) and show superior generalization to DF (95.27\%, 95.60\%) as compared to all other methods. However, average AUC drops to 79.71\% for CAST-B0 and 83.36\% for CAST-B5, primarily due to weak performance on FS (51.33\%) and F2F (74.42\%). This suggests that while NT captures strong textural distortions, its temporal dynamics are less diverse, limiting its usefulness as a generalizable training set. Comparing the two backbones, CAST-B0 consistently outperforms CAST-B5 across most settings. CAST-B0 achieves the best average AUCs when trained on DF (84.65\%) and F2F (83.79\%), while also maintaining better cross-manipulation stability. We attribute this to EfficientNet-B0's reduced parameter count, which may help prevent overfitting and improve generalization under moderate data regimes.

Overall, the results confirm the effectiveness of our model, particularly when trained on DF or F2F data. The model also showcases strong generalization across unseen manipulations when trained on these datasets. The performance trends further highlight the role of manipulation diversity and artifact strength in shaping generalization capabilities of deepfake detection models.

\begin{table}[htbp]
\centering
\caption{Single-manipulation training evaluation on FF++ (HQ) in terms of AUC (\%). Each model is trained on one specific manipulation subset and tested across all four subsets (DF, F2F, FS, NT). Intra-manipulation results are highlighted in gray. EfficientNet-B0 and EfficientNet-B5 as CNN backbone variants are considered for comparison.}
\label{tab:single_manipulation_auc}
\begin{adjustbox}{width=0.8\textwidth}

\begin{tabular}{llcccccc}
\hline
\textbf{Training Set} & \textbf{Method} & \textbf{Year} & \textbf{DF} & \textbf{F2F} & \textbf{FS} & \textbf{NT} & \textbf{Avg} \\
\hline

&  D\textsuperscript{2}-Fusion~\cite{QIU2025103087}  & 2025   & 99.98 & 77.88 & 62.25 & 75.73 & 78.96 \\
& DCL~\cite{Sun_Yao_Chen_Ding_Li_Ji_2022}           & 2022    & 99.98 & 77.13 & 61.01 & 75.01 & 78.28 \\
& RECCE~\cite{Cao_2022_CVPR}                        & 2022    & 99.65 & 70.66 & 74.29 & 67.34 & 70.76 \\
DF
& ADAL~\cite{9931753}                           &  2023      & 99.68 & 79.41 & 73.61 & 81.92 & 83.65 \\
& WATCHER~\cite{Wang2024}                        & 2024       & 99.68 & 75.50 & \textbf{79.51} & 83.66 & 84.58\\
& CAST-B5(Ours)                                    & 2025        & 99.95 & 82.43 & 59.88 & 86.41 & 82.16 \\
& CAST-B0(Ours)                                     & 2025       & \textbf{100.00} & \textbf{86.29} & 60.15 & \textbf{92.02} & \textbf{84.65} \\
\hline

&  D\textsuperscript{2}-Fusion~\cite{QIU2025103087} & 2025    & 89.50 & \textbf{99.86} & 62.47 & 75.23 & 81.76 \\
& DCL~\cite{Sun_Yao_Chen_Ding_Li_Ji_2022}            & 2022   & 91.91 & 99.21 & 59.58 & 66.67 & 79.34 \\
& RECCE~\cite{Cao_2022_CVPR}                        & 2022    & 75.99 & 98.06 & 64.53 & 72.32 & 70.95 \\
F2F
& ADAL~\cite{9931753}                          & 2023         & 90.32 & 99.17 & \textbf{69.49} & 73.13 & 83.03 \\
& WATCHER~\cite{Wang2024}                          & 2024     & 85.73 & 99.82 & 69.38 & 74.16 & 82.27\\
& CAST-B5(Ours)                                     & 2025       & 88.64 & 98.61 & 65.69 & 70.19 & 80.78 \\
& CAST-B0(Ours)                                     & 2025       & \textbf{94.33} & 99.68 & 58.42 & \textbf{82.74} & \textbf{83.79} \\
\hline

&  D\textsuperscript{2}-Fusion~\cite{QIU2025103087} & 2025    & 77.50 & 69.76 & 99.92 & 58.45 & 76.41 \\
& DCL~\cite{Sun_Yao_Chen_Ding_Li_Ji_2022}      & 2022         & 74.80 & 69.75 & 99.90 & 52.60 & 74.26 \\
& RECCE~\cite{Cao_2022_CVPR}                      & 2022      & 82.39 & 64.44 & 98.82 & 56.70 & 67.84 \\
FS
& ADAL~\cite{9931753}                              & 2023     & 72.36 & 70.20 & 99.91 & \textbf{62.12} & 76.14 \\
& WATCHER~\cite{Wang2024}                           & 2024    & \textbf{84.52} & 75.06 & 99.95 & 58.81 & \textbf{79.58} \\
& CAST-B5(Ours)                               & 2025             & 58.13 & 74.71 & \textbf{99.99} & 53.61 & 71.61 \\
& CAST-B0(Ours)                                      & 2025      & 59.97 & \textbf{83.39} & 99.29 & 51.11 & 73.44 \\
\hline

&  D\textsuperscript{2}-Fusion~\cite{QIU2025103087}  & 2025   & 94.44 & 71.08 & \textbf{80.75} & \textbf{99.43} & \textbf{86.42} \\
& DCL~\cite{Sun_Yao_Chen_Ding_Li_Ji_2022}       & 2022        & 91.23 & 79.31 & 52.13 & 98.97 & 80.41 \\
& RECCE~\cite{Cao_2022_CVPR}                        & 2022    & 78.83 & \textbf{80.89} & 63.70 & 93.63 & 74.47 \\
NT
& ADAL~\cite{9931753}                                & 2023   & 90.94 & 63.28 & 78.47 & 99.28 & 82.99 \\
& WATCHER~\cite{Wang2024}                             & 2024  & 89.67 & 72.85 & 69.96 & 98.58 & 82.76 \\
& CAST-B5(Ours)                                        & 2025    & \textbf{95.6} & 77.73 & 62.17 & 97.97 & 83.36 \\
& CAST-B0(Ours)                                        & 2025    & 95.27 & 74.42 & 51.33 & 97.82 & 79.71 \\
\hline
\end{tabular}
\end{adjustbox}
\end{table}

Overall, the results confirm the effectiveness of our model, particularly when trained on DF or F2F data. The model also showcases strong generalization across unseen manipulations when trained on these datasets. The performance trends further highlight the role of manipulation diversity and artifact strength in shaping generalization capabilities of deepfake detection models.

\subsection{Intra-dataset evaluation on FF++}

We evaluate the performance of our proposed models under the intra-dataset setting using the high-quality (HQ) subset of the FF++ dataset. Table~\ref{tab:intra_auc_comparison} compares our EfficientNet-based models (CAST-B0 and CAST-B5) against several SOTA deepfake detection methods in terms of accuracy (ACC) and area under the ROC curve (AUC). Among existing methods, RECCE~\cite{Cao_2022_CVPR}, MAT~\cite{Zhao_2021_CVPR} and D\textsuperscript{2}-Fusion~\cite{QIU2025103087} demonstrate strong performance in terms of AUC (\%). Our proposed CAST-B5 model achieves the highest AUC of 99.49\% as compared to these methods. Although CAST-B5 achieved 97.57\% accuracy which is slightly lower than D\textsuperscript{2}-Fusion who achieved 97.77\%, it maintains a strong accuracy against MAT (97.60\%). Furthermore, CAST-B5 outperforms other notable methods such as F3-Net~\cite{10.1007/978-3-030-58610-2_6} with 97.52\%, GFF~\cite{Luo_2021_CVPR} with 96.87\%, and RECCE with 97.06\%, indicating a clear margin of improvement in accuracy and highlighting the effectiveness of our approach. The CAST-B0 model, though slightly lower in performance, still achieves competitive results with 96.89\% accuracy and 98.98\% AUC. This result highlights its efficacy as a lightweight alternative while retaining robust detection performance. The relatively small performance gap between CAST-B0 and CAST-B5 also suggests that the cross-attention mechanism contributes significantly to generalization and representation power, even when used with a lower-capacity CNN backbone. These results confirm the effectiveness of our CNN-Transformer architecture in capturing both spatial and temporal features for intra-dataset detection. The performance of CAST-B5, demonstrates impressive performance on FF++ (HQ), showing that our cross-attention design effectively bridges spatial cues from individual frames with sequence-level temporal context.

\begin{table}[htbp]
\centering
\caption{Intra-dataset evaluation on FF++ (HQ) in terms of ACC(\%) and AUC (\%) compared with other state-of-the-art methods. Only AUC values under High Quality compression are shown.}
\label{tab:intra_auc_comparison}
\begin{adjustbox}{width=0.5\textwidth}

\begin{tabular}{lccc}
\hline
\textbf{Method} & \textbf{Year} & \textbf{ACC(\%)} & \textbf{AUC(\%)} \\ 
\hline
Face X-ray~\cite{Li_2020_CVPR}       & 2020 & - & 87.35 \\
F3-Net~\cite{10.1007/978-3-030-58610-2_6} & 2020 & 97.52 & 98.10 \\
GFF~\cite{Luo_2021_CVPR}         & 2021 & 96.87 & 98.95 \\
MAT~\cite{Zhao_2021_CVPR}           & 2021 & 97.60 & 99.29 \\
DCL~\cite{Sun_Yao_Chen_Ding_Li_Ji_2022} &2022 & - &99.30\\
RECCE~\cite{Cao_2022_CVPR}          & 2022 & 97.06 & 99.32 \\
D\textsuperscript{2}-Fusion~\cite{QIU2025103087}    & 2024 & \textbf{97.77} & 99.42 \\
CAST-B5(Ours)                        & 2025 & 97.57 &
\textbf{99.49}\\
CAST-B0(Ours)                        & 2025 & 96.89 & 98.98 \\
\hline
\end{tabular}
\end{adjustbox}
\end{table}

\begin{table}[t]
\centering
\caption{Cross-dataset evaluation on Celeb-DF v2, DFD and DFDC by training on FF++ (HQ), compared with other methods in terms of AUC (\%).}
\label{tab:cross_dataset}
\renewcommand{\arraystretch}{1.2}
\begin{adjustbox}{width=0.45\textwidth}
\begin{tabular}{lcccc}
\hline
\textbf{Method} & \textbf{Year} & \multicolumn{3}{c}{\textbf{AUC (\%)}} \\
\cmidrule(lr){3-5}
& & \textbf{DFD} & \textbf{Celeb-DF} & \textbf{DFDC} \\
\hline
Face X-ray~\cite{Li_2020_CVPR}  & 2020 & 85.60 & 74.20 & 70.00 \\
GFF~\cite{Luo_2021_CVPR}              & 2021 & 91.90 & 79.40 & 79.70\\
MAT~\cite{Zhao_2021_CVPR}             & 2021 & - & 67.40 & -\\
LTW~\cite{Sun_Liu_Ye_Gao_Liu_Shao_Ji_2021} & 2021 & 88.56 & 77.14 & 69.00\\
RECCE~\cite{Cao_2022_CVPR}            & 2022 & - & 68.71 & 69.06\\
DCL \cite{Sun_Yao_Chen_Ding_Li_Ji_2022}  & 2022 & 91.66  & \textbf{82.30} & 76.71 \\
SFDG~\cite{Wang_2023_CVPR}               & 2023 & 88.00 & 75.83 & 73.64\\
MeST-Former~\cite{LIU2024128588}  &   2024  &  72.16  &  76.33 & -\\
ID-MAM~\cite{Sheng2025}               & 2025 & - & 72.04 & -\\
CAST-B5 (Ours)                        & 2025 & 92.07  & 74.92 & 78.39\\
CAST-B0 (Ours)                        & 2025 & \textbf{93.31} & 76.98 & \textbf{81.25}\\
\hline
\end{tabular}
\end{adjustbox}
\end{table}

\subsection{Cross-dataset evaluation}

To further validate the generalization ability of our proposed models, we perform cross-dataset evaluation by training on the high-quality (HQ) subset of FaceForensics++ (FF++) and testing on three challenging unseen datasets: Celeb-DF (v2), DeepFakeDetection (DFD) and Deepfake Detection Challenge (DFDC). This evaluation simulates realistic deployment conditions where the model encounters manipulation methods and identities that were not part of its training distribution. Table~\ref{tab:cross_dataset} presents the results in terms of AUC (\%) for both datasets, compared with leading state-of-the-art methods.

Among prior works, DCL~\cite{Sun_Yao_Chen_Ding_Li_Ji_2022} demonstrates the highest overall performance on the Celeb-DF benchmark, achieving an AUC of 82.30\%, followed by GFF~\cite{Luo_2021_CVPR} and LTW~\cite{Sun_Liu_Ye_Gao_Liu_Shao_Ji_2021}, which attain AUC scores of 79.40\% and 77.14\%, respectively. On the DFD dataset, both GFF and DCL exhibit strong performance, reporting AUC scores of 91.90\% and 91.66\% respectively. In contrast, our proposed CAST-B0 model achieves a superior AUC of 93.31\% on DFD, indicating enhanced generalization capabilities compared to prior methods. On the more challenging Celeb-DF benchmark, characterized by high-quality and artifact-free manipulations, CAST-B0 achieves an AUC of 76.98\%. While this performance is slightly below that of DCL, GFF, and LTW, it remains highly competitive. Furthermore, CAST-B0 surpasses other recent methods, including SFDG~\cite{Wang_2023_CVPR} (75.83\%), Face X-ray~\cite{Li_2020_CVPR} (74.20\%) and MeST-Former~\cite{LIU2024128588} (76.33\%). On DFDC, GFF and DCL achieve AUC scores of 79.70\% and 76.71\% respectively, highlighting the increased difficulty of this dataset due to its diverse manipulations and realistic scenarios. Comparatively, our proposed CAST-B0 model achieves a superior AUC of 81.25\% on DFDC, outperforming all prior methods on this benchmark. The more complex CAST-B5 variant also yields competitive results, achieving AUC scores of 92.07\% on DFD, 74.92\% on Celeb-DF and 78.39\% on DFDC. Notably, it outperforms several state-of-the-art baselines such as Face X-ray~\cite{Li_2020_CVPR} (74.20\%), MAT~\cite{Zhao_2021_CVPR} (67.40\%), and ID-MAM~\cite{Sheng2025} (72.04\%).

In summary, the cross-dataset results confirm that our CAST-B0 model outperforms all SOTA methods on DFDC and DFD, whle demonstrating competitive performance on Celeb-DF. The CAST-B5 model achieved slightly lower AUC scores, but still outperformed the majority SOTA methods. These results present an improved generalization across different deepfake generation methods and datasets.

\subsection{Multi-source manipulation evaluation}

To assess the generalization capability of our model across different manipulation types, we perform a multi-source evaluation on the FF++(HQ) dataset. In this setup, the model is trained on three manipulation methods and tested on the remaining unseen method. This cross-manipulation setting allows us to evaluate how well the model generalizes to novel forgery techniques not encountered during training. Table~\ref{tab:multi_source} presents the AUC (\%) performance for several state-of-the-art models on three different manipulation types: Deepfakes (DF), Face2Face (F2F), and NeuralTextures (NT). Our CAST-B0 model demonstrates superior generalization across all three evaluation settings. On the DF split, CAST-B0 achieves an AUC of 97.04\%, outperforming recent models like Implicit~\cite{Huang_2023_CVPR} (95.03\%) and D\textsuperscript{2}-Fusion~\cite{QIU2025103087} (95.34\%). For the F2F subset, CAST-B0 again leads with an AUC of 87.50\%, showing notable improvement over MAT~\cite{Zhao_2021_CVPR} (78.52\%). The most significant gain is observed in the NT split, where CAST-B0 achieves a substantial AUC of 84.50\%, outperforming D\textsuperscript{2}-Fusion by over 4\% and showing a 16.3\% improvement over MAT.

These results highlight the effectiveness of our CAST architecture in learning transferable representations that generalize across different manipulation methods. The consistent gains across all manipulation types highlight the model’s strong domain adaptability.

\begin{table}[htbp]
\centering
\caption{Multi-source manipulation evaluation with other methods on FF++ (HQ) in terms of AUC (\%).}
\label{tab:multi_source}
\renewcommand{\arraystretch}{1.2}
\begin{adjustbox}{width=0.45\textwidth}
\begin{tabular}{lcc}
\hline
\multicolumn{3}{c}{\textbf{DF}} \\
\hline
\textbf{Method} & \textbf{Year} & \textbf{AUC(\%)} \\
\hline
MAT~\cite{Zhao_2021_CVPR}             & 2021 & 85.94 \\
LTW~\cite{Sun_Liu_Ye_Gao_Liu_Shao_Ji_2021}  & 2021 & 92.70 \\
Implicit~\cite{Huang_2023_CVPR}           & 2023 & 95.03 \\
D\textsuperscript{2}-Fusion~\cite{QIU2025103087} & 2024 & 95.34 \\
CAST-B0 (Ours)    & 2025 & \textbf{97.04} \\
\hline
\multicolumn{3}{c}{\textbf{F2F}} \\
\hline
\textbf{Method} & \textbf{Year} & \textbf{AUC(\%)} \\
\hline
MAT~\cite{Zhao_2021_CVPR}             & 2021 & 78.52 \\
LTW~\cite{Sun_Liu_Ye_Gao_Liu_Shao_Ji_2021}  & 2021 & 80.20 \\
Implicit~\cite{Huang_2023_CVPR} & 2023 & 84.37 \\
D\textsuperscript{2}-Fusion~\cite{QIU2025103087} & 2024 & 87.23 \\
CAST-B0 (Ours)    & 2025 & \textbf{87.50} \\
\hline
\multicolumn{3}{c}{\textbf{NT}} \\
\hline
\textbf{Method} & \textbf{Year} & \textbf{AUC(\%)} \\
\hline
MAT~\cite{Zhao_2021_CVPR}             & 2021 & 68.20 \\
LTW~\cite{Sun_Liu_Ye_Gao_Liu_Shao_Ji_2021}  & 2021 & 77.30 \\
Implicit~\cite{Huang_2023_CVPR}         & 2023 & - \\
D\textsuperscript{2}-Fusion~\cite{QIU2025103087} & 2024 & 80.26 \\
CAST-B0 (Ours)    & 2025 & \textbf{84.50} \\
\hline
\end{tabular}
\end{adjustbox}
\end{table}

\begin{table}[t]
\centering
\caption{Ablation study: Comparison of model variants using intra- and cross-dataset AUC (\%) scores.}
\label{tab:ablation-results}
\begin{adjustbox}{width=\textwidth}
\begin{tabular}{lcccc}
\toprule
\textbf{Model Variant} & \multicolumn{4}{c}{\textbf{AUC (\%)}} \\
\cmidrule(lr){2-5}
& \textbf{FF++ (HQ)} & \textbf{Celeb-DF} & \textbf{DFD} & \textbf{DFDC} \\
\midrule
No Cross-Attn (CNN + Transformer)     & 98.12 & 66.18 & 90.10 &  69.26         \\
Self-Attn (Spatial + Temporal sep.)   & 96.38 & 71.41 & 87.24 &  66.75         \\
Cross-Attn reversed (Spatial $\rightarrow$ Temporal) & 97.93 & 71.25 & 90.41 & 71.50 \\
Multi-Scale Integration (CNN Layers)  & 98.04  & 72.33  & 88.64 & 72.06 \\
No projection layer                   & 98.60  &  68.14  & 88.49 & 74.19      \\
CAST-B0         & \textbf{98.98} & \textbf{76.98} & \textbf{93.31} & \textbf{81.25} \\ 
\bottomrule
\end{tabular}
\end{adjustbox}
\end{table}

\subsection{Ablation Studies}

To assess the contribution of each key module in our proposed CAST framework, we conducted a comprehensive ablation study. All experiments are performed using the FF++(HQ) dataset for training, with both intra-dataset evaluation and cross-dataset generalization testing using Celeb-DF, DFD and DFDC.

\subsubsection{Impact of Cross-Attention Mechanism}

To evaluate the influence of the cross-attention mechanism, we constructed a baseline CNN-Transformer model by removing the cross-attention block from the CAST model. In this configuration, temporal features derived from the transformer encoder using frame-wise CNN features were directly used for final classification without explicit spatial-temporal interaction. From results in Table~\ref{tab:ablation-results}, we observe that the absence of cross-attention led to a noticeable drop in cross-dataset generalization, with AUC scores decreasing to 66.18\% on Celeb-DF, 90.10\% on DFD and 69.26\% on DFDC, all substantially lower than the CAST model. Although intra-dataset performance on FF++ remained relatively high (98.12 \%), the model’s limited strength across datasets underscores the necessity of cross-attention in aligning spatial and temporal representations. This mechanism enables the model to capture nuanced manipulation traces, thereby enhancing the consistency and reliability of detection under diverse real-world conditions.

\subsubsection{Evaluating Cross-Attention against Decoupled Attention Fusion}

To assess the contribution of direct spatial-temporal interaction, we replace the proposed cross-attention module with independent self-attention layers applied to the spatial and temporal feature streams, followed by feature fusion via concatenation. As shown in Table~\ref{tab:ablation-results}, this decoupled configuration results in an intra-dataset AUC of 96.38\% on FF++ (HQ), notably lower than the CAST model. More critically, its performance on cross-dataset evaluations dropped to 71.41\% on Celeb-DF, 87.24\% on DFD and 66.75\% on DFDC, indicating diminished generalization. These findings suggest that self-attended spatial and temporal features, when combined without explicit interaction, lack the capacity to model complex manipulation artifacts such as asynchronous facial dynamics or localized geometric inconsistencies. In contrast, cross-attention provides a structured mechanism for aligning spatial and temporal representations, enabling more robust detection across varied data distributions.

\begin{figure}[!t]
\centering
\includegraphics[width=0.78\textwidth]{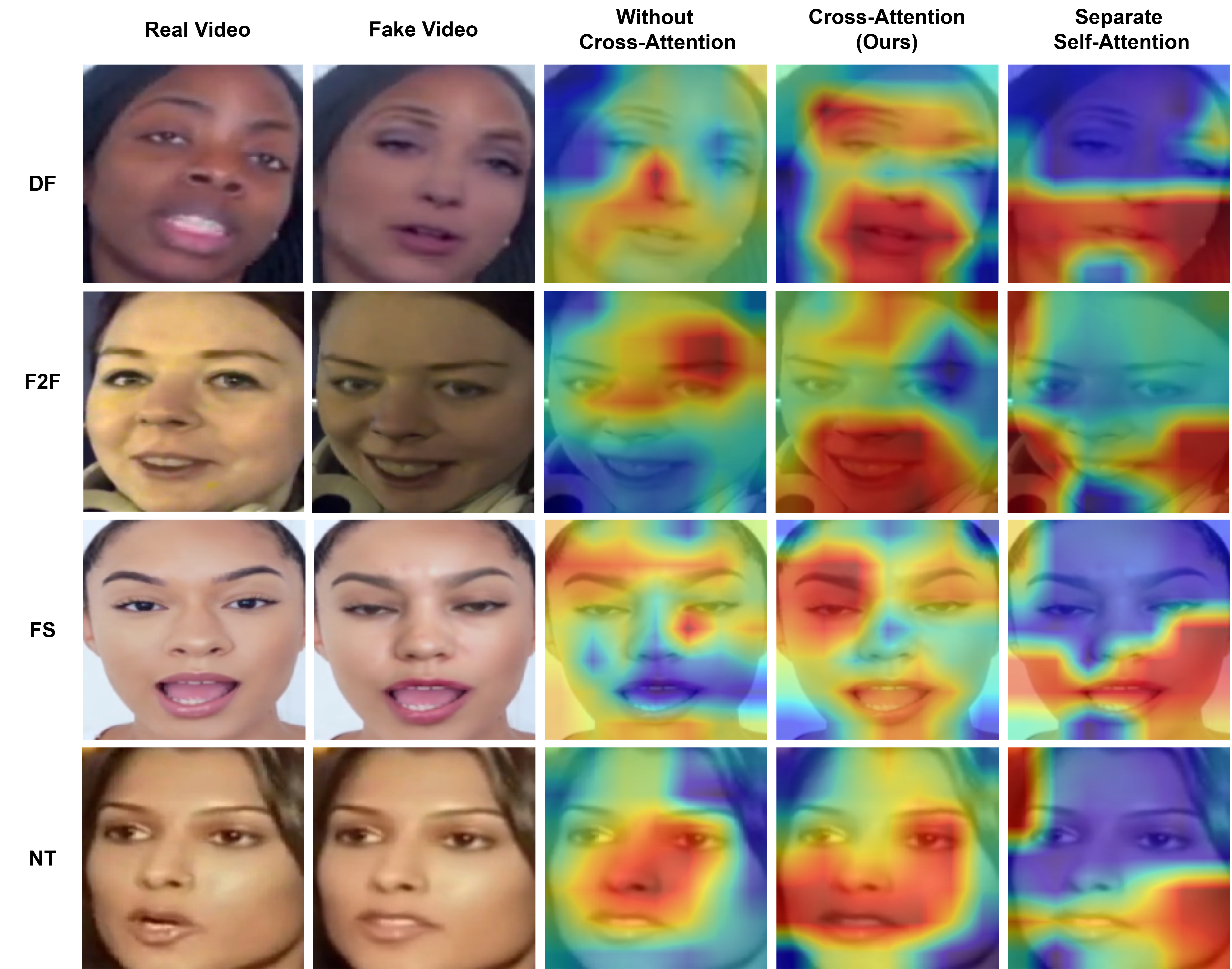}

\vspace{1mm}
\caption{Qualitative analysis on FF++(HQ) manipulation types: Deepfakes (DF), Face2Face (F2F), FaceSwap (FS), and NeuralTextures (NT). From left to right is the real frame, the corresponding fake frame, and the attention heatmap generated by three model variants: without cross-attention, with the proposed cross-attention mechanism, and with decoupled attention fusion.}
\label{fig:ablation_cross_attention_grid}
\end{figure}

Fig.~\ref{fig:ablation_cross_attention_grid} qualitatively supports the ablation results by comparing attention maps across manipulation types. Without cross-attention, the model exhibits diffuse and unfocused activations, while the separate self-attention setup leads to fragmented responses. In contrast, our cross-attention mechanism, where temporal features attend over spatial representations, produces sharper and more localized activations around manipulated regions. This demonstrates its ability to integrate spatial context dynamically, resulting in more interpretable and generalizable representations.

\subsubsection{Effect of Cross-Attention Directionality}

To examine the influence of attention directionality within the cross-attention module, we modified the original configuration by reversing the query and key-value roles by assigning spatial tokens as queries and temporal tokens as key-value inputs. As shown in Table~\ref{tab:ablation-results}, this reversed setup yielded an AUC of 97.93\% on FF++ (HQ), which is marginally lower than the CAST model. However, a more pronounced drop was observed in cross-dataset settings, with performance decreasing to 71.25\% on Celeb-DF and 71.50\% on DFDC, although DFD reduced by a small margin to 90.41\%. These outcomes suggest that leveraging temporal tokens as queries to guide spatial feature refinement is more effective, likely due to temporal dynamics offering a stronger prior for identifying manipulation-relevant spatial regions. This directional design not only improves cross-domain performance but also aids in the interpretability of the model by aligning temporal cues with spatial inconsistencies.

\begin{figure}[!t]
\centering
\hspace*{-0.1\textwidth}
\includegraphics[width=0.55\textwidth]{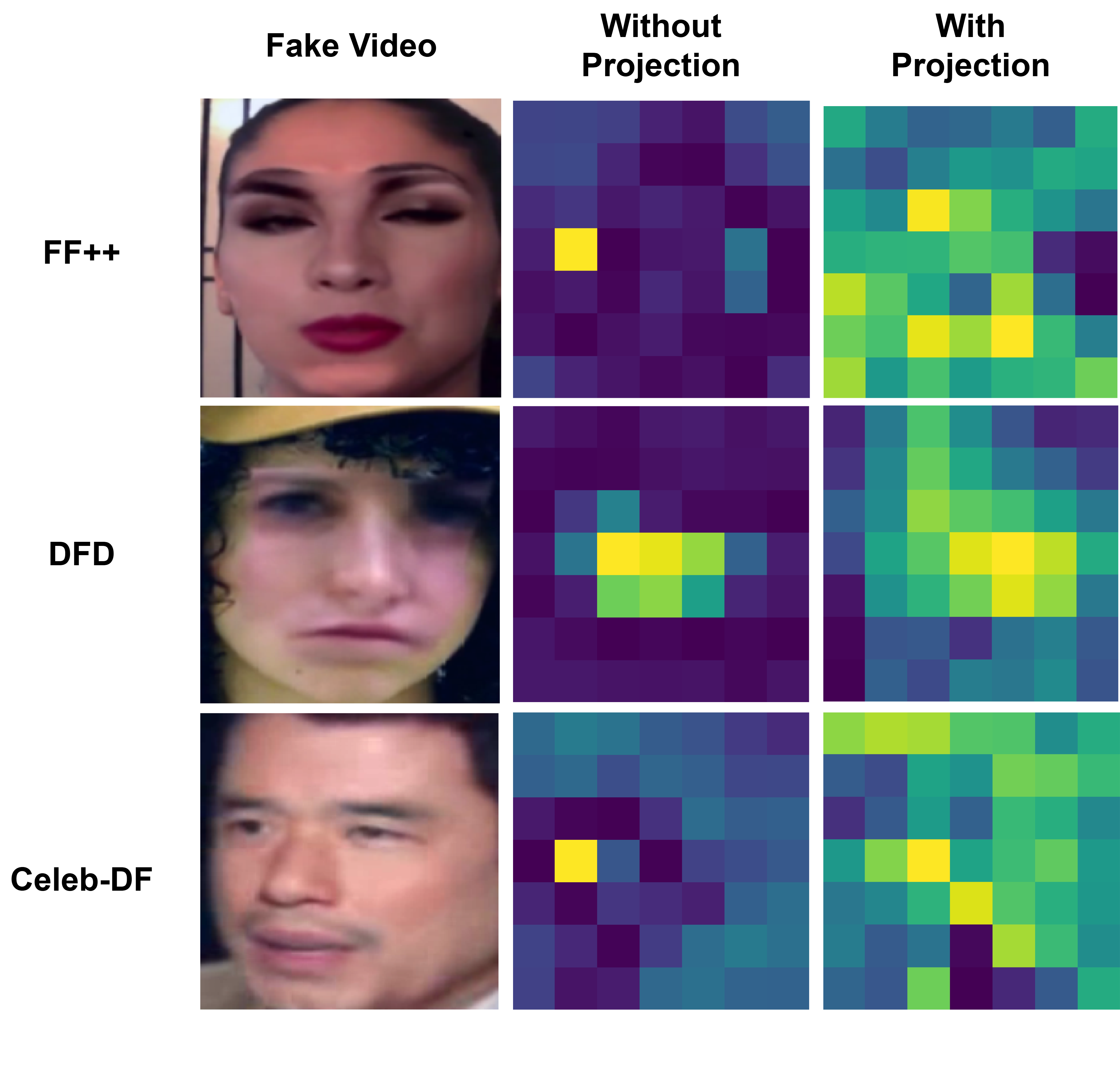}
\vspace{1mm}
\caption{Visual comparison of CNN feature map activations with and without the $1 \times 1$ projection layer across different datasets. Each row corresponds to a sample from a manipulated video (FF++, DFD, and Celeb-DF v2).}
\label{fig:proj_ablation_vis}
\end{figure}

\subsubsection{Effect of Multi-Scale Feature Integration}

To analyze whether incorporating hierarchical spatial representations benefits the CAST architecture, we constructed a multi-scale variant using the same EfficientNet-B0 backbone. Intermediate feature maps from multiple stages of the backbone were extracted, concatenated and projected to the transformer embedding space for temporal modeling. These fused multi-scale embeddings were then passed as the key and value parameters of the cross-attention module, thereby isolating the effect of spatial scale aggregation without altering the backbone family or capacity class.

However, the inclusion of multi-scale features from CNN layers resulted in a consistent decline in performance compared to the baseline CAST-B0 model. The multi-scale variant achieved AUC scores of 98.04\% on FF++ (HQ), 72.33\% on Celeb-DF, 88.64\% on DFD and 72.06\% on DFDC, all notably lower than the corresponding CAST-B0 results. This degradation appears to stem from the representational mismatch introduced by combining heterogeneous low-level and high-level features into a single spatial embedding, which weakens the alignment between spatial frame descriptors and temporally encoded tokens. Since CAST relies on clean, semantically coherent spatial features for effective cross-attention, introducing such heterogeneous mixtures disrupts spatial–temporal correspondence and ultimately hampers the fusion process.

These results indicate that, within the CAST framework, relying on a single high-level semantic representation extracted from the CNN layer provides more stable and discriminative spatial cues than aggregating multi-scale features, reaffirming the importance of consistent spatial embedding quality.

\subsubsection{Convolutional Projection Layer}

To investigate the role of the $1 \times 1$ convolutional projection layer beyond simple dimensionality reduction, we conducted a study on its impact within our CNN-transformer framework. The CAST model already uses EfficientNet-B0 with an output feature dimension of 1280 which is identical to the transformer encoder's embedding size. This setup allowed us to evaluate the effect of the convolutional projection purely in terms of feature adaptation and stability, rather than dimensional compression. We bypass the $1 \times 1$ convolution and directly feed the raw CNN feature maps into the cross-attention module. Similarly, we skipped the convolutional projection after Global Average Pooling (GAP) and directly used the pooled feature maps to prepare for the transformer encoder. From the result reported in Table~\ref{tab:ablation-results}, removing the convolution leads to a noticeable drop in performance, particularly on the more challenging datasets. The AUC on Celeb-DF v2 dropped from 76.98\% to 68.14\%, from 93.31\% to 88.49\% on DFD and from 81.25\% to 74.19\% on DFDC, while FF++ (HQ) maintained relatively high accuracy at 98.60\%. These findings highlight that, even when the dimensionalities are aligned, the convolutional projection serves a crucial role in refining feature representations and ensuring a stable attention mechanism.

To better understand this effect, Fig.~\ref{fig:proj_ablation_vis} illustrates how the CNN feature maps differ when the projection layer is present versus when it is removed. The feature maps without projection tend to appear sparse or contain high-contrast isolated activations, indicating that critical semantic patterns may not be effectively captured or preserved. In contrast, the projected feature maps demonstrate richer, more spatially coherent activations, highlighting their ability to encode more meaningful and task-relevant facial cues. These visual patterns align with our quantitative results and validate the importance of a convolutional projection layer for the attention mechanisms involved in the architecture.

\subsubsection{Impact of CNN Backbone Architecture}

To assess the effect of backbone architecture on spatial feature quality and downstream classification, we evaluated three CNN extractors: EfficientNet-B0, EfficientNet-B5, and XceptionNet. As summarized in Table~\ref{tab:cnn-backbone}, EfficientNet-B5 achieved the best overall AUC score of 99.49\% on FF++(HQ). However, EfficientNet-B0 slightly outperformed it on the Celeb-DF, DFD and DFDC dataset, achieving an AUC of 76.98\%, 93.31\% and 81.25\% respectively, suggesting superior generalization under domain shift. XceptionNet, while commonly used in earlier works, yielded the lowest performance across all datasets.

\begin{figure}[!t]
\centering
\includegraphics[width=\textwidth]{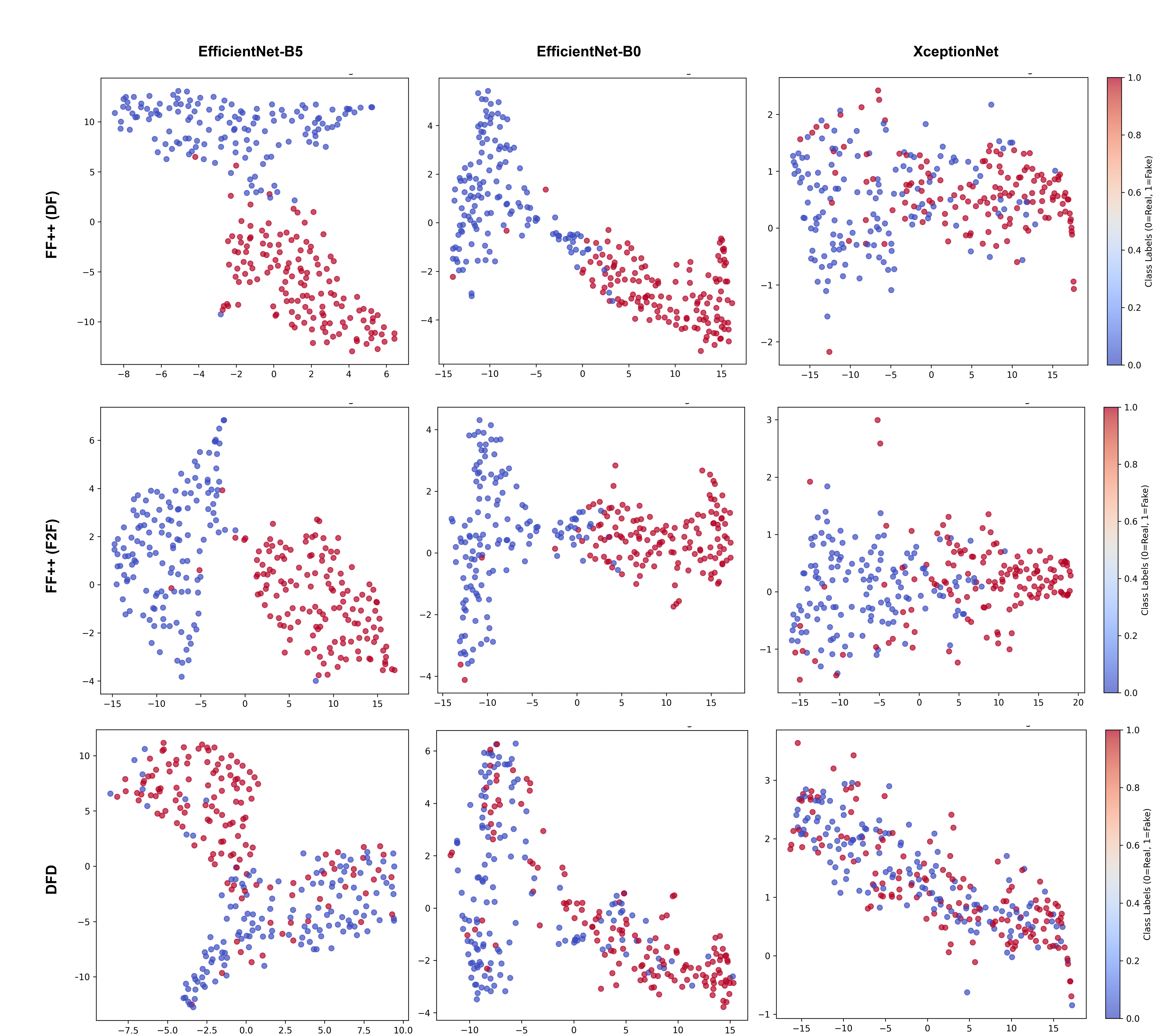}
\vspace{1mm}
\caption{t-SNE visualization of fused feature embeddings across three CNN backbones: EfficientNet-B5, EfficientNet-B0, and XceptionNet on three benchmark datasets: FF++ (Deepfakes and Face2Face subsets) and DFD. Red and blue denote fake and real samples respectively.}
\label{fig:ablation_cnn_backbone}
\end{figure}

To further understand these differences, we visualized the high-level feature representations from each backbone using t-SNE, as shown in Fig.~\ref{fig:ablation_cnn_backbone}. Each subplot presents the separability of real and fake samples for a given dataset–backbone pair. EfficientNet-B5 consistently produces well-separated clusters for both FF++ dataset (DF and F2F), with minimal overlap between real and manipulated distributions. EfficientNet-B0 shows slightly more dispersion yet maintains clear boundaries. In contrast, XceptionNet’s embeddings are noticeably entangled, especially on DFD, confirming its relatively weaker spatial encoding. These patterns highlight the importance of selecting a backbone that balances spatial discriminability with performance across manipulation types and data domains.

\begin{table}[!t]
\centering
\caption{Impact of CNN backbone on model performance. Comparison of model variants using AUC scores based on intra-dataset testing on 
FF++(HQ) and cross-dataset testing on Celeb-DF, DFD and DFDC datasets. All models are variants of our proposed CAST model with different CNN backbones.}
\label{tab:cnn-backbone}
\begin{adjustbox}{width=0.8\textwidth}
\begin{tabular}{lccccc}
\toprule
\textbf{CNN Backbone} & \multicolumn{4}{c}{\textbf{AUC (\%)}} & \textbf{Year} \\
\cmidrule(lr){2-5}
& \textbf{FF++ (HQ)} & \textbf{Celeb-DF} & \textbf{DFD} & \textbf{DFDC} \\
\midrule
EfficientNet-B0~\cite{pmlr-v97-tan19a} & 98.98 & \textbf{76.98} & \textbf{93.31} & \textbf{81.25} & 2019\\
EfficientNet-B5~\cite{pmlr-v97-tan19a} & \textbf{99.49} & 74.92 & 92.07 & 78.39 & 2019\\
Xception~\cite{Chollet_2017_CVPR}      & 97.82 & 66.04 & 89.10 & 60.63 & 2017\\
\bottomrule
\end{tabular}
\end{adjustbox}
\end{table}

\section{Limitations}

While the proposed CAST framework demonstrates strong performance in both intra- and cross-dataset evaluation, several limitations remain that merit attention.

First, in Table~\ref{tab:single_manipulation_auc}, we observe that although the model performs competitively in intra-manipulation settings achieving AUCs of 100.00\% (DF) and 99.29\% (FS), and achieves high AUC scores in majority of the cross-testing scenarios as well, its generalization across manipulation types is comparatively degraded in case of the FS sub-dataset. Notably, when trained solely on FS, CAST-B0 achieves only 59.97\% and 51.11\% AUC on DF and NT respectively. Similar trends are observed for CAST-B5, with 58.13\% and 53.61\% AUC on those subsets. These results indicate that representations learned from FS are highly specialized and exhibit limited transferability. This challenge extends to cross-testing conditions as well. When trained on DF, the CAST-B0 model’s AUC on FS drops to just 60.15\%, as compared to 79.51\% of WATCHER~\cite{Wang2024}, 74.29\% of RECCE~\cite{Cao_2022_CVPR} and 73.61\% of ADAL~\cite{9931753}, highlighting its difficulty in detecting face-swapping manipulations not seen during training. 
This limitation can be attributed to the characteristics of the FaceSwap manipulation in FF++, where artifacts are relatively subtle and exhibit limited temporal variation across frames~\cite{roessler2019faceforensicspp}. Unlike DF and F2F, which manipulate every frame of the target sequence, FS modifies only a subset of overlapping frames, resulting in weaker and less consistently distributed temporal cues. Since CAST relies on temporal--spatial inconsistencies to guide cross-attention fusion, the reduced strength of temporal signals in FS limits the contribution of the temporal branch and degrades cross-manipulation generalization.

Second, from Table~\ref{tab:intra_auc_comparison}, we note that although our models achieve competitive AUC values (98.98\% for B0 and 99.49\% for B5) on FF++ (HQ), the classification accuracy (ACC) lags slightly behind other state-of-the-art approaches. Specifically, CAST-B0 achieves 96.89\% ACC, the lowest among listed methods, while CAST-B5 reaches 97.57\%, still underperforming compared to MAT (97.60\%). This discrepancy between AUC and ACC suggests that although the models rank predictions effectively, their decision thresholds may not be ideally calibrated under high-quality compression, possibly due to the subtle nature of discriminative features in such settings.

A similar limitation is observed in the cross-dataset evaluation on Celeb-DF, where CAST achieves competitive yet comparatively lower performance than on datasets such as DFD. Celeb-DF consists of high-fidelity, artifact-sparse deepfake videos with explicitly reduced temporal flickering through the use of temporally correlated facial landmarks, resulting in smoother frame-to-frame transitions and fewer temporal inconsistencies~\cite{Celeb_DF_cvpr20}. Consequently, the reduced availability of temporal irregularities limits the effectiveness of detection models that rely on temporal cues. This challenge is further amplified by the significant distribution shift between FF++ and Celeb-DF.

These performance limitations can partly be attributed to computational constraints. Due to hardware limitations, all experiments are conducted using 16 frames, sampled at a regular interval, from the video clips. While this frame count was chosen to balance temporal coverage and memory usage, it was likely unable to explore the model’s ability to capture longer temporal inconsistencies.

\section{Conclusion}

In this study we presents CAST, a novel Cross-Attentive Spatio-Temporal feature fusion architecture tailored for robust deepfake video detection. Specifically, we introduce a cross-attention mechanism to effectively fuse spatial and temporal features extracted from a CNN-Transformer network. By explicitly modeling the interplay between spatial and temporal features, the proposed cross-attention mechanism enables temporal representations to attend selectively to manipulation-prone spatial regions such as those affected by unnatural warping, expression inconsistencies, or flickering artifacts. Hence,
enhancing the network’s capacity to localize and detect subtle forgeries.
Comprehensive evaluations are conducted across multiple benchmarks, including FaceForensics++ (HQ), Celeb-DF v2, DFD and DFDC under a variety of testing protocols, including intra-dataset, cross-manipulation, and multi-source setups. CAST consistently outperformed and was comparable to SOTA methods in most settings. Specifically, under the high-quality single-manipulation scenario, CAST-B0 variant achieved 100\% AUC on DF, 86.29\% on F2F and 92.02\% on NT, while CAST-B5 achieved 99.99\% on FS, confirming the framework’s strong detection capabilities when trained on
single manipulations. In intra-dataset evaluation, CAST-B5 achieved the highest AUC score of 99.49\% as compared to the SOTA methods along with a comparable accuracy of 97.57\%. In cross-manipulation and multi-source contexts, CAST demonstrated favorable generalization, outperforming SOTA approaches when cross-tested on DFD and DFDC datasets, achieving 93.31\% and 81.25\% AUC scores respectively, while maintaining competitive performance when evaluated on unseen Celeb-DF videos. Ablation studies confirmed the critical role of cross-attention in enhancing generalization, with its removal leading to significant drops in cross-dataset performance. Replacing it with decoupled self-attention further underscored the importance of explicit spatial-temporal interaction. Directionality experiments showed that using temporal queries are more effective for aligning manipulation cues. Moreover, the convolutional projection layer is essential for stable attention computation and backbone comparisons reveal that CAST-B5 achieves higher intra-dataset accuracy whereas CAST-B0 shows better generalization, making it well-suited for efficient real-world deployment.

The proposed model CAST provides a principle and extensible approach to deepfake detection through cross-modal feature fusion. Future directions include extending CAST to handle audio-visual deepfakes by integrating audio modality cues with the existing spatial-temporal framework. Additionally, optimizing CAST for real-time deployment on edge devices remains a promising direction. We also aim to investigate scaling of the model to larger temporal contexts.

\bibliographystyle{elsarticle-num}

\bibliography{bibliography}




\end{document}